\documentclass[10pt,twocolumn,letterpaper]{article}
\pdfoutput=1
\usepackage{cvpr}
\usepackage{times}
\usepackage{epsfig}
\usepackage{graphicx}
\usepackage{amsmath}
\usepackage{amssymb}
\usepackage{xparse}

\usepackage[pagebackref=true,breaklinks=true,letterpaper=true,colorlinks,bookmarks=false]{hyperref}

\cvprfinalcopy % *** Uncomment this line for the final submission

\begin{document}

%%%%%%%%% TITLE
\title{StableNet: Semi-Online, Multi-Scale Deep Video Stabilization}

\author{Chia-Hung Huang$^1$\footnotemark[2]
\quad
Hang Yin$^1$\footnotemark[2]
\quad
Yu-Wing Tai$^2$
\quad
Chi-Keung Tang$^1$\\
 \\
$^1$Hong Kong University of Science and Technology\\
$^2$Tencent Youtu\\
{\tt\small \{chuangag, hyinac\}@connect.ust.hk, yuwingtai@tencent.com, cktang@cs.ust.hk}
}

\maketitle
\footnotetext[2]{Equal contribution.}
\thispagestyle{empty}

%%%%%%%%% ABSTRACT
\begin{abstract}
    Video stabilization algorithms are of greater importance nowadays with the prevalence of hand-held devices which unavoidably produce videos with undesirable shaky motions. In this paper we propose a data-driven online video stabilization method along with a paired dataset for deep learning. The network processes each unsteady frame progressively in a multi-scale manner, from low resolution to high resolution, and then outputs an affine transformation to stabilize the frame. Different from conventional methods which require explicit feature tracking or optical flow estimation, the underlying stabilization process is learned implicitly from the training data, and the stabilization process can be done online. Since there are limited public video stabilization datasets available, we synthesized unstable videos with different extent of shake that simulate real-life camera movement. Experiments show that our method is able to outperform other stabilization methods in several unstable samples while remaining comparable in general. Also, our method is tested on complex contents and found robust enough to dampen these samples to some extent even it was not explicitly trained in the contents.
\end{abstract}

%%%%%%%%% BODY TEXT
\section{Introduction}

    Videos captured by hand-held cameras and mobile phones usually contain undesirable shaky motions, which may lead to hardship of perceiving the contents. Digital video stabilization is a post-process on a given shaky video that eliminates the high frequency jitters and retains a stable and visually acceptable result. An efficient and effective video stabilization algorithm would be beneficial due to the prevailing possession of smart phones that serve as video recording devices and are prone to shaky hand motions. With a successful hardware-independent video stabilization system, users could produce preferable videos without purchasing expensive physical stabilizers.

    Most prior video stabilization algorithms followed the steps of ``camera path estimation" $\rightarrow$ ``camera path smoothing" $\rightarrow$ ``stabilized frame synthesis based on smoothed path", which require offline path estimations that are based on feature detection or optical flow
    \begin{figure}[t]
    \begin{center}
    \includegraphics[scale=0.27]{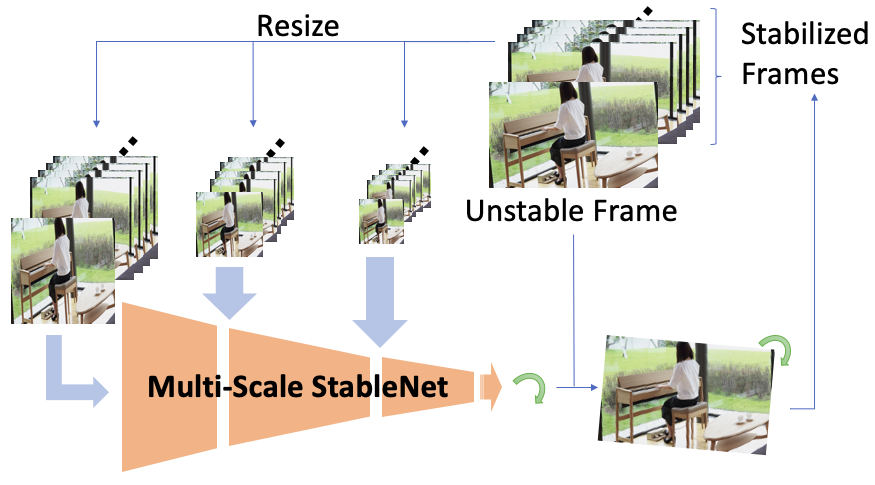}
    \end{center}
       \caption{Overview of Multi-Scale StableNet. The input is a stack of historical stable frames with an unstable frame. We resize the input into 3 levels and process them in sequence. The final output is an affine transform that stabilizes the unstable frame. Please see our supplemental materials which include all input and output videos.}
    \label{fig:proposed_model}
    \label{fig:onecol}
    \end{figure}
    estimations and may introduce errors. Different from previous works, we achieve video stabilization by a data-driven and end-to-end approach, without intermediate camera path estimation and explicit feature detection. Also, it is possible for our method to conduct online processing on the video, which is a valuable feature for real-time filming devices.

    With the rapid growth of computational power and availability of large-scale data, data-driven deep learning algorithms such as ConvNet (Convolution Neural Networks) have achieved outstanding performance on other video processing tasks such as video generation~\cite{tulyakov2017mocogan}, deblurring~\cite{su2017deep}, and stylization~\cite{zhu2017exemplar}. In comparison to conventional approaches on video stabilization, data-driven approach requires few explicit manipulations to the optimization process, and is more result-oriented where the loss function can be viewed as the resulting performance. However, to apply deep learning to video stabilization, there exist two main challenges. First, since we are not explicitly estimating and optimizing the camera path, which is a more intuitive approach of stabilization, the stabilization process definition need to be accommodated to be suitable for ConvNet solutions. Second, different from extant algorithms, which were mostly independent from input data, deep learning approaches demands a large amount of training data. The data are not only required to be sufficiently large but also to contain adequate variance.

    To resolve the lack of training data, we synthesized the data based on an intuitive physical method. With proper parameter setting, we are able to generate realistic shaky videos with different extents of jitter. The advantages of using synthesized data include: accurate ground truth output for learning and extensible data contents. As a result, we generated about 420 shaky videos based on about 140 stable video clips, containing over 170000 frames for training.

    For the problem definition, we adopt an online stabilization approach along with multi-scale refinement, as shown in Figure~\ref{fig:proposed_model}. The input to the network are 3 stacks of frames in different resolutions. Each stack contains one unstable frame that is to be stabilized, together with stable historical frames. The historical stable frames are sampled with different time intervals according to the resolution level. Generally, a higher resolution stack is formed by a smaller sampling interval. This setting is used to incorporate different range of temporal information into the network. The network will eventually output an affine transformation to stabilize the unstable frame. In addition to the original online stabilization problem setting, we also proposed a semi-online stabilization method, which first stabilizes part of the video (i.e., a video chunk), then merges the newly stabilized chunk back to the video. This method mitigates the error propagation during online stabilization and enables the network to tackle low-frequency tilts.

    We evaluate our method on two major aspects quantitatively: fidelity and stability. Results show that our system can perform quantitatively better than other online stabilization methods, and even offline stabilization methods in some samples. We also evaluated the robustness of our model on complex contents such as zooming camera and parallax scenes and found our method can still handle these samples to some extent. Our contributions can be summarized as follows: (1) We propose a novel video stabilization algorithm based on deep neural networks and multi-scaling techniques. (2) We propose a synthetic paired training dataset for future video stabilization works.

%-------------------------------------------------------------------------
\begin{figure*}
\begin{center}
\includegraphics[scale=0.45]{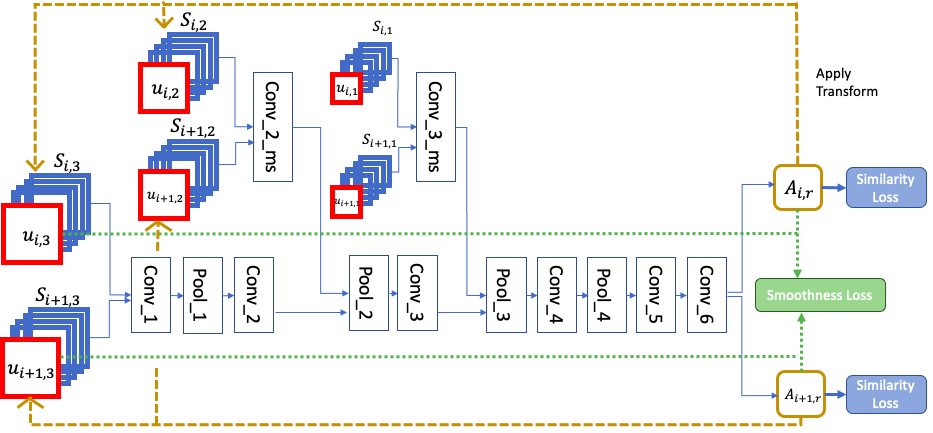}
\end{center}
   \caption{Network Architecture. The Multi-scale StableNet is based on Siamese architecture. Two consecutive unstable frames will be selected and form two input stacks respectively. The frame stacks will then be resized into 3 levels. Two genre of losses are computed separately and back propagated through the network.}
\label{fig:detail_architecture}
\end{figure*}
%-------------------------------------------------------------------------

%------------------------------------------------------------------------
\section{Related Work}
    Conventional video stabilization is performed in three steps as estimating the camera motion trajectory, smoothing the original path and synthesizing the stabilized video sequence with the smooth path to remove undesired motions. The approaches are applied mostly in 2D or 3D perspectives in terms of path estimation, while Jin \etal~\cite{jin2001digital} proposed 2.5D inter-frame motion model and dealt with affine transformation achieving real-time stabilization.

    In 2D camera perspectives, Hu \etal~\cite{hu2007video} extracted scale invariant features and removed the shaky motions by Gaussian kernel filtering. The affine model was also used in~\cite{litvin2003probabilistic} to describe interframe transformation, and recursive Kalman filtering was applied to stabilize the trajectory. Particle filtering improved the estimation in~\cite{yang2009robust} and showed increasing robustness. Grundmann \etal~\cite{grundmann2011auto} presented the algorithm applying L1-optimization with several constraints, and the synthesized smooth path is composed of either constant, linear or parabolic motions. Subspace constraints were implemented in~\cite{liu2011subspace} and applicable to long videos, and generated visually plausible videos. Specifically for the estimation of camera motion, many approaches have been developed. Matsushita \etal~\cite{matsushita2006full} divided the trajectory into global and local motions, and employed motion inpainting to improve video quality. Another method for estimating interframe camera motions is dynamic time warping technique developed by Bosco \etal~\cite{bosco2008digital} intended for moving-object videos. Liu \etal~\cite{liu2014steadyflow} proposed SteadyFlow model dealing with spatially-variant motions.  Camera motions were mitigated under the turbulent conditions by particle advection framework \cite{oreifej2013simultaneous}. Instead of focusing on the camera motion path, Lee \etal~\cite{lee2009video} optimized the feature trajectories of the unstable videos, and produced less undefined space in the stabilized videos.

    As for 3D perspectives which include the focal length variation, Liu \etal~\cite{liu2009content} reconstructed dynamic scenes in 3D camera motion for perceptually plausible results. Goldstein and Fattal~\cite{goldstein2012video} adopted epipolar point transfer for moving camera and object videos matching the capabilities of 3D methods without scene reconstruction. Zhang \etal~\cite{zhang2009video} solved the problem by tackling a sparse linear system of equations. Additionally, Liu \etal~\cite{liu2013bundled} handled camera motions with a bundle of camera paths, and succeeded in dealing with parallax and rolling shutter effects.

    Besides the previous offline approaches, several online methods are also proposed, which are favored by real-time applications. Ratakonda \cite{ratakonda1998real} proposed a primitive online stabilization algorithm of integrating cumulative motion curve in two dimensions separately. Optical flows were calculated between successive frames in \cite{chang2004robust}, and the camera path was optimized with regularization. Liu \etal~\cite{liu2016meshflow} presented MeshFlow using only the past motions. Karpenko \etal~\cite{karpenko2011digital} demonstrated real-time stabilization with gyroscopes for automatic calibration on a mobile phone. Integrating ConvNet, Wang \etal~\cite{wang2018deep} recently proposed StabNet to perform online stabilization by loading historical frames into the network and directly producing the homography matrix entries. However, this method may lead to serious error propagation in that the former lapses cause latter abnormality in terms of the visual performance.

    Our proposed system employs ConvNet to extract the interframe transformation, which does not explicitly estimate the camera motion path or feature trajectory but leave them for the network to learn implicitly. In order to achieve multiple levels of refinement on the stabilization process, we referred to the pyramidal implementation in~\cite{bouguet2001pyramidal}, which improved the performance of optical flow detection, and constructed three levels of different resolutions to do iterative adjustment in stabilization. To our knowledge, our method StableNet is the first which uses deep ConvNet in a multiscale setting for video stablization and produces competitive results compared with state-of-the-art methods.

%------------------------------------------------------------------------
\begin{figure}

      \label{tbl:detail_stat}
      \includegraphics[width=\linewidth]{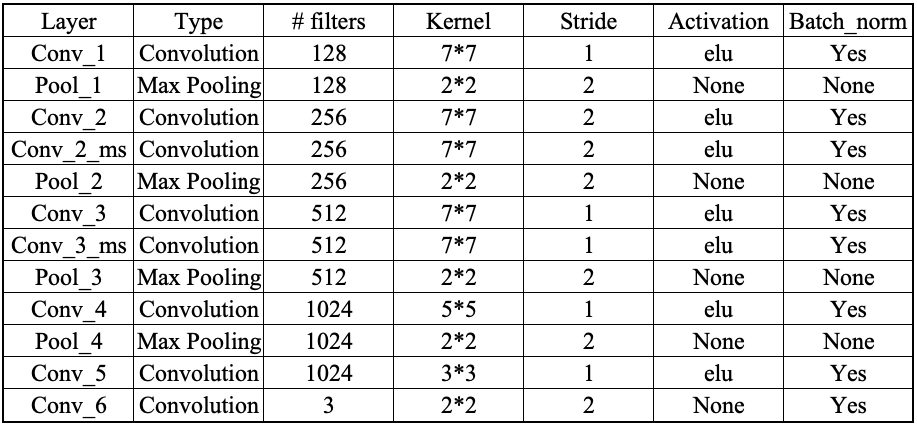}
      \caption{Implementation Details. All padding are in VALID mode. Note that there is no fully connected layer to better preserved the spacial information.}
    \end{figure}
\section{Methodology}
    We approached our task in an online stabilization setting, where the input contains only historical but no future information. The model processes one frame at a time, given stacks $(S, u)$ of historical stabilized frames ($S$) and the current unstable frame ($u$). After the network has captured the latent camera dynamic from the frame stacks, it outputs the corresponding affine transformation to stabilize the current frame. We simplify the affine transformation into 3 parameters $(\theta, dx, dy)$ that are more physically meaningful, corresponding to rotation angle, x-axis translation and y-axis translation. Scaling is excluded since it is considered not to be the major source of instability. For the input of the model, stacks of frames in 3 different resolutions are presented, which leads to a 3-level multi-scale refinement (Section~\ref{net_arch}).  To better incorporate long term and short term interframe dynamics, we set different sampling interval for stacks with different resolutions (Section~\ref{var_temp_sampling}). The model is then trained on the dataset with both supervised and unsupervised loss functions (Section~\ref{loss_func}).
    During testing time, the network stabilizes the video frame by frame. To further mitigate the error propagation caused by the frame by frame stabilization, we propose a chunk-by-chunk stabilization method, which can be considered as a semi-online approach (Section~\ref{semi_online}).
    \subsection{Multi-Scale StableNet Architecture} \label{net_arch}
    We adopt a siamese network architecture as the backbone, followed by a 3-scale setting. All of the layers are convolution layers for better preservation of spacial information. Figure~\ref{fig:detail_architecture} shows the detail of our model. Siamese architecture is chosen since it is proven to excel at capturing interframe dynamics~\cite{Lei_2018_CVPR} which is the critical information for video stabilization. During the training time, $(S_i, u_i)$ and $(S_{i+1}, u_{i+1})$ are passed into the two branches of the network respectively, each containing 3 different resolutions. The different resolution stacks are resized from the original frames. We denote $i\in[1,N]$ as the frame index and $r\in[1,3]$ as the resolution index, where $N$ is the total number of frames. The network processes the lower resolution stack $(S_i, u_i)_r$ and generates an intermediate affine transformation $A_{i_r}$, which will be applied to the next resolution's unstable frame $u_{i_r}$. The network then processes the stack of next resolution level $(S_i, A_{i_r}(u_i))_{r+1}$. The output of each subsequent level serves as a refinement for the previous level.

    \subsection{Variant Temporal Sampling} \label{var_temp_sampling}
    The input stack contains a stack of historical stable frames $S$ and one current unstable frame $u$. The historical frames stack $S$ is constructed by 23 stable frames, sampled with a constant time interval $t$. Thus $S_i=(s_{i-23t}, s_{i-22t}, \cdots, s_{i-t})$, and the first frame is used if the subscript is less than $1$. Given a single $t$, a few problems need to be resolved. If the sampling interval is too large, it will result in incoherent transformation since the input is lacking the dynamics from nearby frames. For sampling interval that is insufficient, long term tilting throughout the video will occur.

    Thus we select different sampling rates for stacks that have different resolutions to better capture temporal information. Larger time interval $t$ is selected for lower resolution stack so the lower resolution input contains long term temporal information. Based on the long term dynamic captured from the low resolution input, the network then outputs the raw affine transformation which will be further refined in next resolution. Since higher resolution stacks serve as small range refinements, the sampling intervals should be smaller, which better provides the information of nearby frames and leads to smooth transformations between frames.
    \subsection{Loss Function} \label{loss_func}
    The loss function consists of two terms, similarity and smoothness:
    \begin{multline}
    L = \sum_{k\in {i, i+1}}{L_{\mathit{similarity}}(A_k, \hat{A_k}, u_k, s_k)} + \\
    \lambda L_{\mathit{smoothness}}(A_i, A_{i+1}, u_{k}, u_{k+1})
    \end{multline}
    where $u_i$ and $s_i$ denote the $i^{th}$ unstable frame and stable frame respectively. $A_i$ is the predicted affine transformation for $u_i$ and $\hat{A_i}$ is the ground truth affine transformation. For simplicity, $A_i$ only contains rotation and translation, thus can be represented as $(\theta_{i}, dx_{i}, dy_i)$. $\lambda$ is the factor for balancing the two terms.
    \subsubsection*{Similarity Loss}
    The similarity loss is a supervised loss, which evaluates the model's predictions against the ground truths. The ground truth affine transformations can be retrieved from the synthetic processes of training data. There are two components which form the similarity loss, one is parameter-wise comparison and the other is image-wise comparison.\\

    The parameter-wise term serves as a stronger guidance for the model output and lead to a faster convergence in the training process. The image-wise term provides a visual quality guidance to the model, which compares the stabilized frame with the ground truth stable frame. The loss can be expressed as
    \begin{multline}
        L_{\mathit{similarity}}(A_k, \hat{A_k}, u_k, s_k) = MSE(A_k, \hat{A_k}) + \\ \dfrac{\alpha}{H*W*D}MSE(A_k(u_k), s_k)
    \end{multline}
    where $H$, $W$, $D$ are the spacial dimension of the image and $\alpha$ is the factor for balancing the terms.
    \subsubsection*{Smoothness Loss}
    The smoothness loss is an unsupervised loss, which evaluate the smoothness between consecutive frames. To measure the smoothness, we first calculate the transform $T_i$ between the two stable frames $(s_i, s_{i+1})$. $T_i$ is estimated based on optical flow calculated from the Lucas–Kanade method~\cite{bouguet2001pyramidal} and Good Features to Track~\cite{Shi94goodfeatures}. Note that this estimation is only needed for calculating training loss, which is not required during testing time. We then measure the difference between $T_i(A_i(u_i))$ and $A_{i+1}(u_{i+1})$, which should be close to zero if both frame are stabilized. The term can be represented as
    \begin{multline}
        L_{\mathit{smoothness}}(A_i, A_{i+1}, u_{k}, u_{k+1}) = \\
        MSE(T_i(A_i(u_i)), A_{i+1}(u_{i+1}))
    \end{multline}
    The losses for the 3 resolution levels are accumulated and back propagated after the last resolution is processed. Note that except for the lowest resolution, the affine transform used in loss calculation is the transform after refinement.
    \begin{figure*}[t]
    \begin{center}
    \includegraphics[scale=0.3]{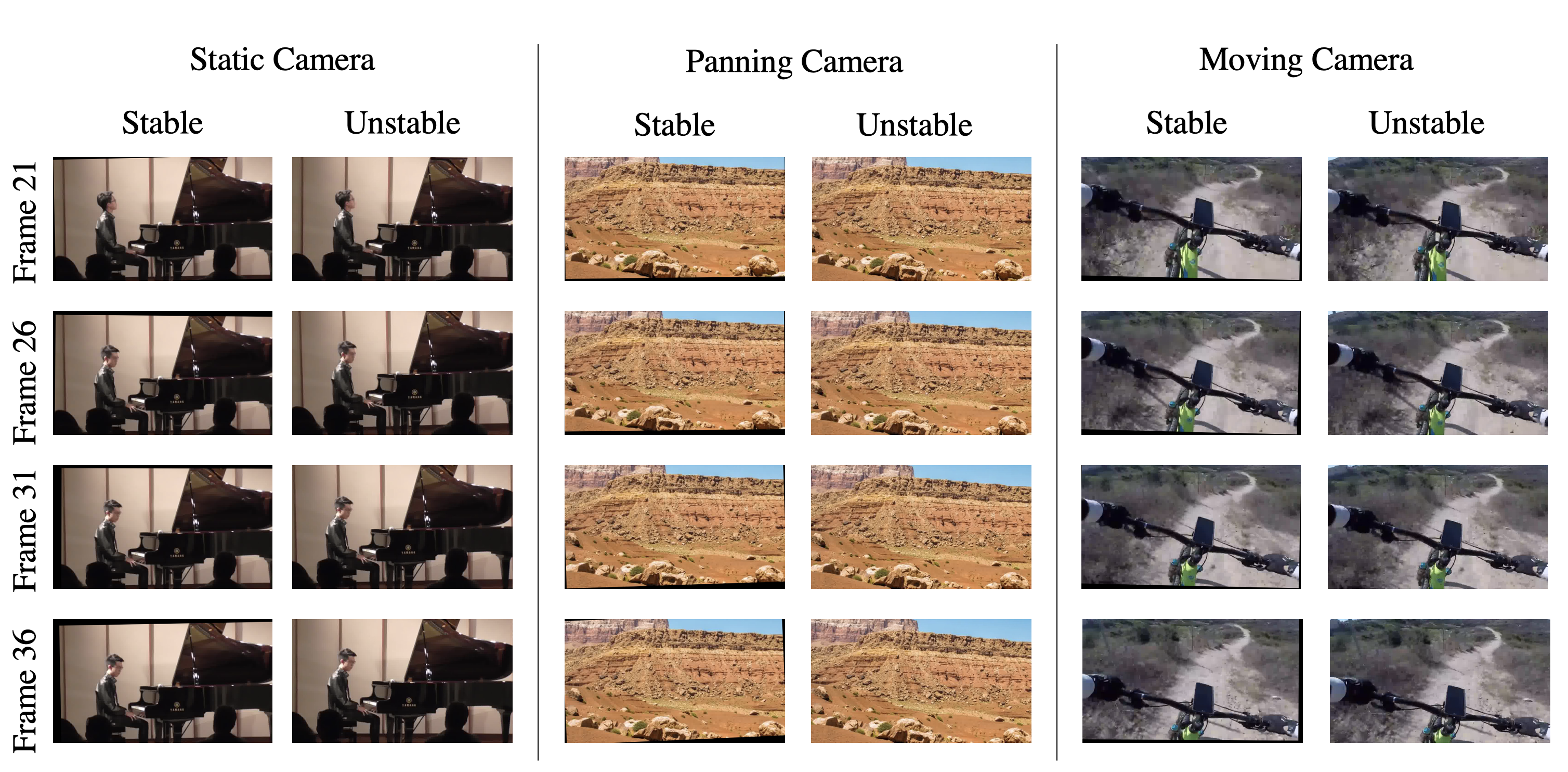}
    %\fbox{\rule{0pt}{2in} \rule{.9\linewidth}{0pt}}
    \end{center}
       \caption{Frames from our dataset. The dataset consists of about 420 pairs of steady and synthesized shaky videos with three extents of camera motions from 140 stable video clips. Black border is persistent in training data to better simulate the input in testing time.}
    \label{fig:dataset}
    \end{figure*}
    \subsection{Semi-online video stabilization} \label{semi_online}
    Ground truth stable frames are stacked as the input during training time. There will be no ground truth stable frames during testing time, so the stack is formed by previously stabilized frames. However, this is prone to error propagation during testing time. For example, the stabilized scene may be incorrectly tilted as the length of the video increased, though the video can still be considered stable. In addition to adopting variant temporal sampling, we also propose a semi-online stabilization method to resolve the error propagation. First, the whole video is divided into small chunks, $(c_1, c_2, \cdots, c_k)$, where each chunk contains a relatively small number (32) of frames, thus $c_i=(u_{32(i-1)+1}, u_{32(i-1)+2}, \cdots, u_{32i})$. The model then stabilizes each chunk individually, assuming the first frame of each chunk to be stable, and results in a series of stabilized chunks $(c'_1, c'_2, \cdots, c'_k)$. As for the chunk merging, each chunk will be applied another chunk-wise affine transform which is based on previously merged chunks and the first frame of the current chunk. This step serves as a chunk-wise stabilization. This method can be interpreted as semi-online since it processes a small batch of frames at a time but not one frame at a time. Nevertheless, it still leverages the advantage of online stabilization due to the small number of frames within each chunk. Furthermore, this methodology successfully reduces the effect of error propagation caused by standard online video stabilization.

    \subsection{Implementation Details}
    The detailed implementation of our network can be viewed in Figure~\ref{tbl:detail_stat}. We trained our network using Adam Solver~\cite{DBLP:journals/corr/KingmaB14} with learning rate of 0.001, an exponential decay of $\gamma=0.98$ for every 5 epochs, and a batch size of 96. It took about 60 hours to converge on about 170000 samples when trained on 4 Nvidia GTX 1080Ti graphic cards.

    We also normalized both the input and the output domain. Observing that the model output was too large in the initial stage of training which resulted in translation that falls out of the borders, we scaled up the target value by 1000 to avoid decimal errors. So each of $(\theta, dx, dy)$ falls within $[-1000, 1000]$, and will be scaled back when constructing the affine matrix. The parameter setting in our experiments is: $\lambda = \alpha =10000$. And we resized the original frame with inter area interpolation to the three resolutions: $30 \times 30$, $125 \times 125$, and $256 \times 256$, which respectively correspond to sampling interval of $6$, $3$, $1$ (frames).

   The stable frames in the input stack are generated by applying ground truth affine transform to unstable frames. This will result in frames with some black borders, which better simulate the input during testing time. Also we have added stable stacks to the training data, thus the correct output should be an identical affine transform. This is to dampen the model from invariably performing transformation, even when there is no such demand. The ratio of stable over unstable training samples is 0.2 in our experiments, which added adequate amount of regularization to the model.
%------------------------------------------------------------------------

\section{Training Data}

    Data collection is one of the challenges we are faced with when dealing with neural network training process. Since there exist few resources for the appropriate datasets including pairs of stable and unstable videos, we constructed our own dataset by applying artificial noise jitters. We have considered the parallel hardware method of collecting training data proposed by Wang \etal~\cite{wang2018deep}, however, errors in the estimation from stable frame to unstable frame may inevitably occur with optical flow detection due to inappropriate feature point selection and other factors, which can cause inconsistency when calculating loss. Therefore, we instead develop a shake synthesis algorithm that includes rotation and translation. Since our proposed method focuses on the stabilization in 2D spaces, we do not include depth changes in video synthesis. In this way, we only need to gather the stable videos as the basis for synthesis, and not only can we achieve synchronization between pairs of videos, but it also manifests efficiency because we are able to extend the ground truth videos to different extent of artificial movement and vibration by adopting various sets of parameters.

    As for stable videos, we selected such one-take videos which do not contain scene transition and subtitles which can affect the content of frame stacks used for training. In order to prevent the network from overfitting, we need to guarantee variety in terms of the camera motion type as well as the video contents. Therefore, the collected videos contain movie clips, music videos, live concerts, sports recordings and so on, and the camera motions generally involve static, panning and locomotive types.

    After the accumulation of about 140 stable videos, each of which is around 15 seconds long at 24 FPS in general, we added the artificial noise on the stable frames by affine transformation matrix derived from rotation angle and translation vector. The matrix $M_t$ is calculated as:
    \[
    M_t=
    \begin{bmatrix}
    \cos\theta & \sin\theta & \Bar{x}(1-\cos\theta)-\Bar{y}\sin\theta+dx\\
    -\sin\theta & \cos\theta & \Bar{x}\sin\theta+\Bar{y}(1-\cos\theta)+dy
    \end{bmatrix}
    \]
    where $\theta$ is the rotation angle, $dx, dy$ are the horizontal and vertical translation, $\Bar{x}=r_x-dx, \Bar{y}=r_y-dy$ with  $r_x, r_y$ as the rotation center (set as 640, 360 in our synthesis since it is the center of the original stable frame).

    We developed three sets of parameters corresponding to small, medium and large extent of shake, and the parameters vary in the range of $\theta, dx, dy$ as well as the time interval between two random assigned parameter. To elaborate the latter, we ensure the synthesized videos to resemble the authentic shaky videos from hand-held devices by exploiting linear interpolation. We generate two random numbers for each parameter within a certain interval (e.g.\ 4, 6 frames), and assign the interpolated value to each of the interim frames according to the index of the frame in the current interval, and in this way the synthesized video will not suffer from sudden and slack changes between adjacent frames. The extent of shake embodies in the initialization of the random number generation, where the standard deviation varies.

    Furthermore, since the stabilized frames are utilized as part of the frame stack in the stabilization process, black borders generated from affine transformation cannot be avoided. We must therefore include the undefined areas into the ground truth videos instead of directly regarding the original as the standard. In order to add plausible borders, we produced a stable video for each synthesize shaky one by transforming the unstable frames with the inverse matrix, which can also be derived from the artificial parameters. In total, we finally have 420 pairs of videos, and they are separated into 390 training pairs and 30 validation pairs, and the validation pairs are carefully selected based on the camera motion types so that it is representative of various stabilization effect. Figure~\ref{fig:dataset} presents some exemplar frames of the dataset.

\section{Experiments and Results}
    In this section, we compared our method against several out-standing video stabilization algorithms~\cite{liu2013bundled,goldstein2012video,grundmann2011auto} and stabilizer of iMovie. The evaluation was done in two major aspects, in terms of the fidelity and stability of the results (Section~\ref{metrics}). The result of the evaluation then will be shown and discussed in Section~\ref{results}. We have also tested on samples that cannot be stabilized perfectly with affine transform only, and found that our model still can somehow stabilize these samples (Section~\ref{robustness}). Finally, current limitations of our method will be examined in Section~\ref{limitation}.
    \subsection{Metrics} \label{metrics}
    \subsubsection*{Fidelity}
    As suggested by Morimoto \etal~\cite{eval1998}, the fidelity of video stabilization system can be evaluated using the peak signal-to-noise ratio (PSNR).
    \[
    \mathit{PSNR}(I_0,I_1)=10\log\dfrac{255^2}{\mathit{MSE}(I_0,I_1)}
    \]
    Intuitively, a fully stabilized image sequence will have no residual motion in ideal case. Thus there will be no difference between pixels of two consecutive frames. Thus the larger the PSNR between two consecutive frames, the more stable is the sequence. While the objects in the scene usually move between the transition of frames, which will cause difference in pixel values even the video is stabilized, this intuitive evaluation metric still provides an insight of the performance of the stabilization system. In our experiment, we measured the average PSNR across the entire stabilized video as the fidelity of the method.
    \begin{figure*}
    \begin{center}
    \includegraphics[scale=0.52]{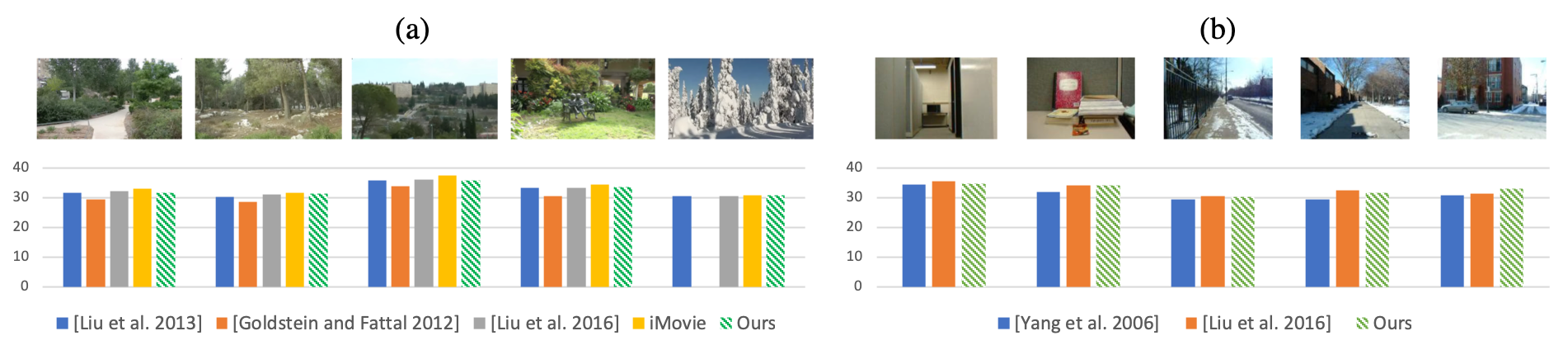}
    \end{center}
       \caption{Fidelity experiments results. The fidelity is measured by calculating the average interframe PSNR (in dB): (a) shows the evaluation against general state of the art methods and commercial software, (b) shows the comparison against online stabilization methods specifically. Results show that our method is comparable to state of the art methods and slightly outperforms other online methods.}
    \label{fig:fidelity_result}
    \end{figure*}
    \begin{figure*}
    \begin{center}
    \includegraphics[scale=0.55]{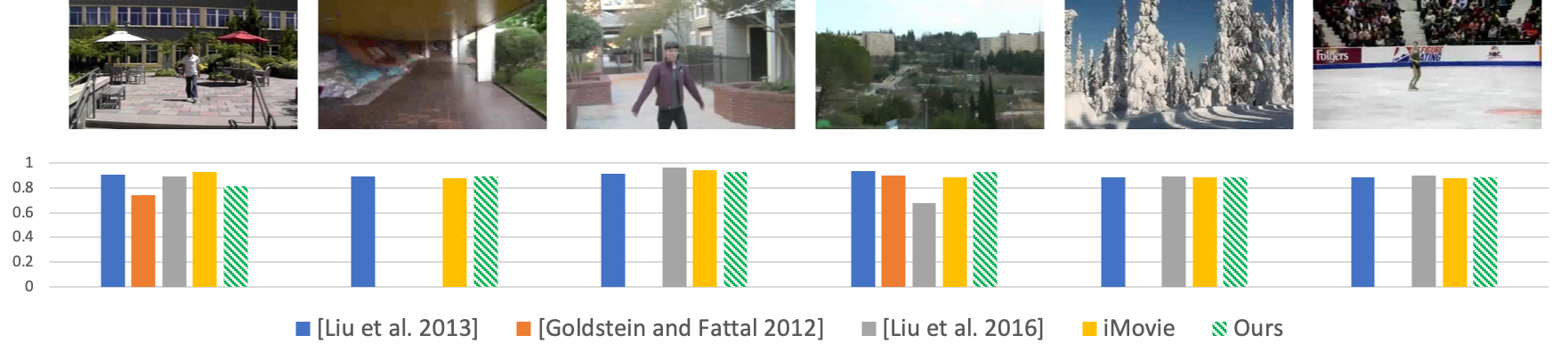}
    \end{center}
       \caption{Stability experiments results. Stability is measured based on the minimum energy percentage in rotation, horizontal translation and vertical translation by frequency analysis. The comparison is made among previous stabilization algorithms, representative of commercial product, iMovie, and our proposed model StableNet.}
    \label{fig:stability_result}
    \end{figure*}

    \subsubsection*{Stability}
    Following the criteria proposed by \cite{liu2013bundled}, we measured the stability of result videos by a frequency domain analysis on 2D motions of the estimated camera path. The camera path is estimated as accumulative Homography transformations between frames: $P^t=H_0H_1..H_t$, where the interframe homography is calculated from feature-based optical flow detection. A larger power proportion of low frequency motions represents a more stable path. In ideal case of stabilization, the proportion should close to 1. In our experiment, we calculated the ratio of the power sum of the 2nd to the 7th lowest frequency components against the total power of all frequency components. We evaluated the energy percentages of rotation, horizontal translation and vertical translation components and selected the minimum values among these three values as the stability metric of the video.
    \begin{figure*}
      \begin{center}
      \includegraphics[width=\linewidth]{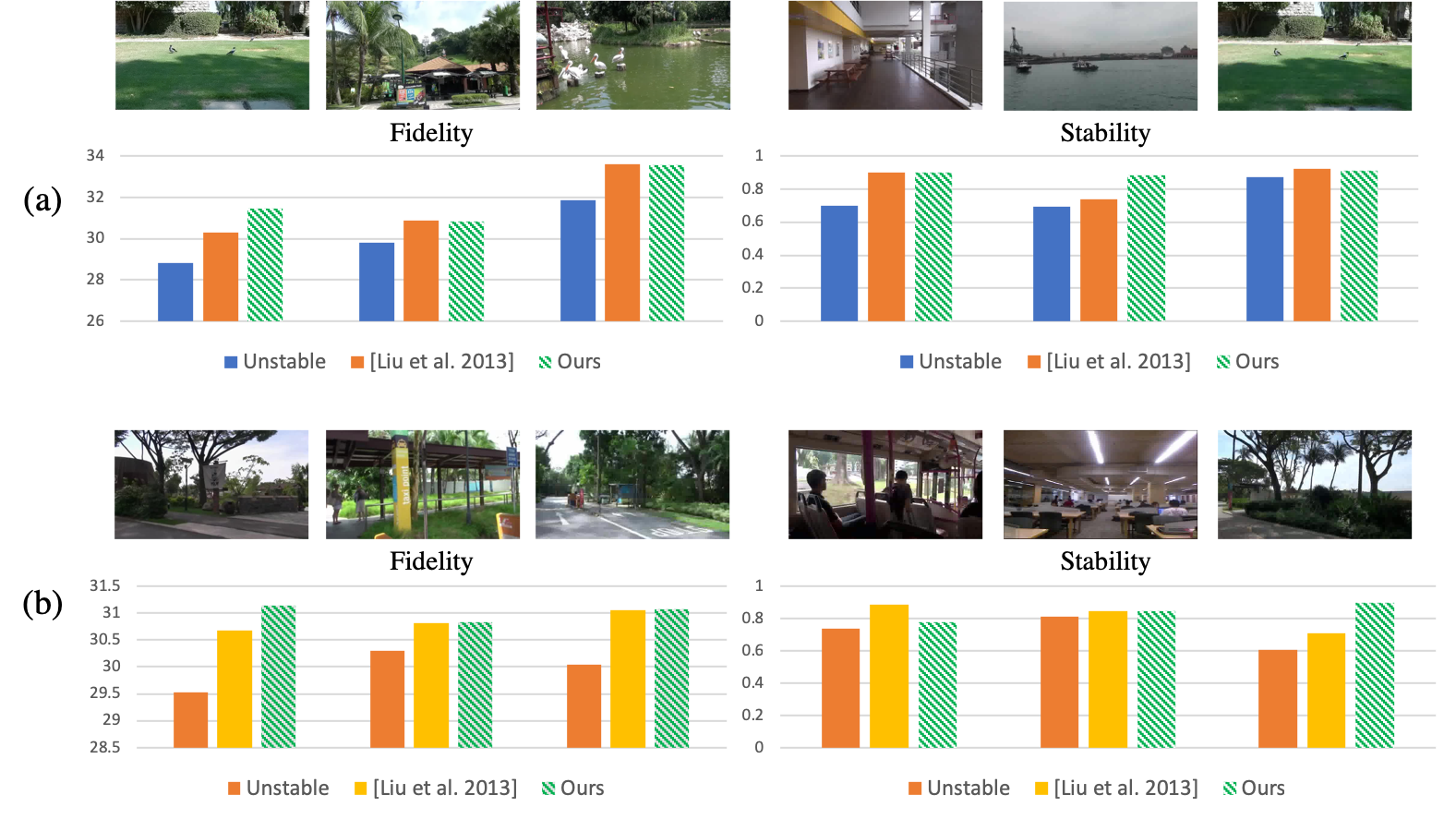}
      \end{center}
      \caption{Fidelity and stability for (a) zooming and (b) parallax videos. Although there is no scaling or grid warping in the output affine transformation for our model, our model is shown to be robust enough to handle zooming and parallax scenes. This could be attributed to the using of dataset with large variance, which contains these more complex scenes. }
      \label{fig:zooming_parallax}
    \end{figure*}
    \subsection{Results and Evaluation} \label{results}
    The experiments were conducted on video clips presented in \cite{liu2013bundled}. We compared works from \cite{liu2013bundled,goldstein2012video,grundmann2011auto}, and also the stabilizer of iMovie. In terms of the video fidelity, it is shown in Figure~\ref{fig:fidelity_result} (a) that our method can sometimes outperform other offline, path optimization based methods and being comparable to commercial software. As for the stability measurement, it performs similarly to the fidelity where our approach sometimes rivaling other offline algorithms according to Figure~\ref{fig:stability_result}. The result that it cannot constantly outperform other offline methods was expected since our method is an online method. The lacking of future information and the error propagation during stabilization will penalize the performance.

    Figure~\ref{fig:fidelity_result} (b) shows the comparison against other online methods on fidelity. The comparison is done on the data published in \cite{particlefilter2006}. The result shows that our method outperform \cite{particlefilter2006} and have comparable results to \cite{liu2016meshflow} in the fidelity metrics. As for the stability compared with online methods, Figure~\ref{fig:stability_result} shows comparable result between our method and Liu \etal~\cite{liu2016meshflow}.

    \subsection{Robustness} \label{robustness}
    Notice that though our model targets in solving regular jitters which are caused by walking and unintended hand movements, our method is robust enough to stabilize more complex contents to some extent, such as zooming camera and indoor scenes with parallax. The evaluation is shown in Figure~\ref{fig:zooming_parallax}, which is done on the Zooming and Parallax category of videos from \cite{liu2013bundled}. The result shows that our model is quantitatively comparable to \cite{liu2013bundled} on this type of contents even the settings did not explicitly handle them.

    This could be a merit of using a synthetic dataset that contains these type of contents implicitly. Though our dataset is synthesized with simple transformation only, the scalibility on both the contents and jitter patterns of synthetic dataset is an advantage that can offer additional variance. Our large dataset works well with deep neural networks and leads to robustness.

    \subsection{Limitation} \label{limitation}
    While our method is able to handle regular shaky videos that contain simple 2D motions, it is restricted from handling other more complex problems such as handling roller shutter effects by nature due to the relatively simple output. This could be a future work of modifying the model output to better handle more complex interframe motions. Another limitation of our method is that it handles high frequency jitters better than low frequency jitters, as well as small motions against large motions, which may lead to some results that seem to have low visual stability in these cases. This may be a result of the limited underlying patterns in our training data, which did not contain drastic motions. Adding more shaking patterns in the training data should be able to solve this issue.

%------------------------------------------------------------------------
\section{Conclusion}
    We have presented StableNet, our semi-online multi-scale stabilization approach with deep learning techniques. The proposed model can stabilize unstable frames by extracting and analyzing latent patterns in historical frames. Our method is free from motion estimation and can be applied to online stabilization. We have also proposed a paired dataset for future data-driven video stabilization works, which contains about 420 synthesized videos of three jitter types based on public videos. Quantitative evaluations showed that our model is not only able to perform stabilization comparably with previous state of the arts and commercial software on regular shaky videos but have also attained robustness from the using of our large synthetic dataset.

{\small
\bibliographystyle{ieee}
\bibliography{final}
}

\end{document}